\documentclass[journal]{ieeetran}
\usepackage[colorlinks=true,
allcolors=green]{hyperref}
\usepackage{hyperref}
\usepackage{url}
\usepackage{tablefootnote}
\usepackage[noadjust]{cite}
\usepackage{amsmath,amssymb,amsfonts}
\usepackage{algorithmic}
\usepackage{graphicx}
\usepackage{textcomp}
\usepackage{subcaption}
\usepackage{adjustbox}
\usepackage{caption}
\usepackage{algorithm2e}
\usepackage{makecell}
\usepackage{multirow}
\usepackage{xcolor}
\usepackage{threeparttable}
\usepackage{verbatim}
\RestyleAlgo{ruled}
\usepackage{mathrsfs}
\usepackage{bm}
\graphicspath{ {./images} }
\SetKwComment{Comment}{/* }{ */}
\newcommand\norm[1]{\lVert#1\rVert}

\begin{document}
\title{Analysis of Fatigue-Induced Compensatory Movements in Bicep Curls: Gaining Insights for the Deployment of Wearable Sensors}
\author{Ming Xuan Chua, Yoshiro Okubo, Shuhua Peng, Thanh Nho Do, Chun Hui Wang, and~Liao~Wu,~\IEEEmembership{Member,~IEEE}
\thanks{This work involved human subjects. Approval of the ethical, experimental procedures and protocols was granted by The University of New South Wales under the HC reference number: HC220573.} 
\thanks{This work was supported by UNSW Engineering International Collaboration Seed Grant PS70042.}
\thanks{Corresponding author: Liao Wu}
\thanks{Ming Xuan Chua is with the School of Mechanical and Manufacturing Engineering, The University of New South Wales (UNSW) and Falls, Balance and Injury Research Centre, Neuroscience Research Australia (NeuRA) (email:mingxuan.chua@unsw.edu.au; c.chua@neura.edu.au)}
\thanks{Yoshiro Okubo is with the Falls, Balance, and Injury Research Centre, Neuroscience Research Australia, Australia (email: y.okubo@neura.edu.au).}
\thanks{Shuhua Peng, Chun Hui Wang, and Liao Wu are with the School of Mechanical and Manufacturing Engineering, The University of New South Wales, Australia (email: shuhua.peng@unsw.edu.au; chun.h.wang@unsw.edu.au; liao.wu@unsw.edu.au).}
\thanks{Thanh Nho Do is with the Graduate School of Biomedical Engineering, The University of New South Wales, Australia (email: tn.do@unsw.edu.au).}
}
\maketitle

\begin{abstract}

A common challenge in Bicep Curls rehabilitation is muscle compensation, where patients adopt alternative movement patterns when the primary muscle group cannot act due to injury or fatigue, significantly decreasing the effectiveness of rehabilitation efforts. The problem is exacerbated by the growing trend toward transitioning from in-clinic to home-based rehabilitation, where constant monitoring and correction by physiotherapists are limited. Developing wearable sensors capable of detecting muscle compensation becomes crucial to address this challenge. This study aims to gain insights into the optimal deployment of wearable sensors through a comprehensive study of muscle compensation in Bicep Curls. We collect upper limb joint kinematics and surface electromyography signals (sEMG) from eight muscles in 12 healthy subjects during standard and fatigue stages. Two muscle synergies are derived from sEMG signals and are analyzed comprehensively along with joint kinematics. Our findings reveal a shift in the relative contribution of forearm muscles to shoulder muscles, accompanied by a significant increase in activation amplitude for both synergies. Additionally, more pronounced movement was observed at the shoulder joint during fatigue. These results suggest focusing on the shoulder muscle activities and joint motions when deploying wearable sensors to effectively detect compensatory movements.

\end{abstract}


\label{sec:introduction}
\section{Introduction}
\IEEEPARstart{B}{icep} Curl exercise is commonly prescribed for rehabilitating upper limb diseases such as bicep tendonitis \cite{churgay2009diagnosis}, an inflammation of the bicep tendon caused by repetitive overhead motion or age-related wear, among other reasons. The treatment usually consists of bicep curl exercises with high repetition and low weight. It strengthens the Bicep Brachii muscle after inflammation is relieved. The whole rehabilitation comprises four steps: 1) managing pain; 2) increasing flexibility and joint range of motion (ROM); 3) regaining muscle strength; 4) sport-specific functional training \cite{physiopedia2022rehab}. Rehabilitation from bicep tendonitis can be a long course of treatment, requiring patients to undergo supervised procedures in a hospital or clinic first, usually during the acute stage, and then continue the exercise at home for an extended period until fully recovered.

Muscle compensation is a prominent issue that frequently occurs in rehabilitation, including rehabilitating upper limbs involving bicep curls. This terminology describes the behaviors of patients who, instead of using the targeted muscles to complete specific movements, would unconsciously choose other muscles to perform the instructed movements due to muscle fatigue or pain. Consequently, the effectiveness of rehabilitation is significantly degraded and can cause further injury. If rehabilitation treatment lasts for an extended period of time, incorrect muscle use may even lead to permanent postural deviations \cite{levin2009motor}.  

When patients receive rehabilitation treatment in a hospital or clinic, physiotherapists can identify and correct muscle compensation through their direct observation and instruction. However, detecting muscle compensation when patients perform rehabilitation at home is challenging since the process is unsupervised. The problem becomes increasingly severe along with the trend of shifting from in-clinic rehabilitation to home-based rehabilitation due to the dramatically increasing demand from the aging population. Various investigations show that continuity of support outside the hospital is limited \cite{van2004continuity}. Moksnes et al. \cite{moksnes2023factors} also demonstrated the decreasing centrality of the patient's residence has an increasing likelihood of sustaining severe injuries. 

Recent advances in technologies such as virtual reality \cite{lin2022vr}, telehealth \cite{arntz2023technologies}, wearable sensors \cite{wang2022technology}, robotic arms \cite{nicholson2020multi}, and exoskeletons \cite{liu2022home, darmanian2023completely} have relieved the problems faced. Specifically, the advancement of wearable sensors has provided an alternative solution to this problem. While little work has been reported to identify muscle compensation in bicep curls, some sensors have been proposed for similar purposes in other motions, such as reaching objects \cite{cai2020real}, walking \cite{Durandau}, sitting \cite{zaltieri2022assessment}, and sit-to-stand \cite{he2023analysis}. In general, these sensors can be categorized into two groups. The first group captures patient's physical parameters relevant to human movements. Typical examples in this category include marker-free cameras \cite{sugai2023lstm}, motion capture systems \cite{zhao2019biomechanical}, inertial measurement units \cite{nguyen2021quantification,zhao2023analysis,oubre2020simple}, strain sensors \cite{eizentals2020smart}, and pressure sensors \cite{cai2019automatic,cai2020real}. The second group is predominantly based on surface electromyography (sEMG) and detects patient's muscle activities directly \cite{liu2021muscle,afzal2022evaluation,asemi2022handwritten,hajiloo2020effects,ghislieri2020muscle,kubota2020usefulness,liu2022joint,sheng2021metric,yang2017muscle}.

Muscle synergies analysis has been widely used to compare muscle coordination between different stages. It is mainly extracted via a data reduction algorithm, Non Negative Matrix Factorization (NNMF). Dupuis et al. \cite{dupuis2021fatigue} used muscle synergy analysis from sEMG sensors to investigate the changes in shoulder movement during a reaching task after fatigue. Hajiloo et al. \cite{hajiloo2020effects} studied changes in muscle synergy patterns of lower limb muscles during running before and after fatigue and found significant differences in relative weighting and activation of various muscles. Thomas et al. \cite{thomas2023effects} presented a similar work on identifying muscle synergies in the shoulder joint moving a handle. Matsunaga et al. \cite{matsunaga2021muscle} also found that fatigue in the gluteus maximus requires the hamstring to compensate during single-leg landing. Those studies demonstrated that muscle synergy analysis could reflect the changes in the subject's condition, including fatigue. However, none of these studies have investigated the compensatory movements in bicep curls. These works have provided significant features of muscle contribution that can differentiate their condition during exercise. However, limited research connects the changes in muscle synergy patterns with sensor development.

While Bicep Curls primarily comprise the \textit{elbow flexion-extension} motion, its compensatory movements possibly involve \textit{shoulder elevation-depression}, \textit{shoulder flexion-extension}, and \textit{wrist flexion-extension} \cite{baechle2008essentials,kennedy2014methods}. 
During this process, various muscles could be activated, including:
\begin{itemize}
    \item the primary mover muscle: \textit{Biceps Brachii (BIC)};
    \item the synergist muscles: \textit{Brachialis (BR)} and \textit{Brachioradialis (BRA)};
    \item the stabilizer muscles: \textit{Anterior Deltoid (AD)}, \textit{Posterior Deltoid (PD)}, \textit{Upper Trapezius (UT)}, \textit{Flexor Carpi Ulnaris (FCU)}, and \textit{Flexor Carpi Radialis (FCR)}; and
    \item the antagonist muscle: \textit{Triceps (TRI)}.
\end{itemize}

Consequently, to capture and obtain a comprehensive analysis of the Bicep Curl exercise, an array of mechanical and sEMG sensors are required to observe all the joints and muscles stated. It is foreseeable that the wearable sensors developed will be complex and cost-ineffective \cite{vourganas2019factors}. Therefore, it is necessary to conduct a systematic analysis of the compensatory movements in bicep curls to reduce the dimensionality of the features between standard Bicep Curl exercise and fatigue Bicep Curl exercise, to simplify the sensor design and deployment, which has been absent in the existing literature, to the authors' best knowledge. 

To this end, in this work, we use a motion capture system to capture the shoulder and arm joint kinematics such as \textit{shoulder elevation-depression}, \textit{shoulder flexion-extension}, \textit{elbow flexion-extension}, and \textit{wrist flexion-extension} under 
standard and fatigued bicep curls. While the motion capture system is used for analysis here, we note that the results also apply to other mechanical sensors that can measure the joint motions. In addition, we use sEMG sensors to measure eight muscles, including \textit{BIC}, \textit{BRA}, \textit{UT}, \textit{TRI}, \textit{FCU}, \textit{FCR}, \textit{AD}, and \textit{PD}. The \textit{BR} muscle is excluded as it is not a superficial muscle and, therefore, cannot be measured by sEMG appropriately. The rationale for selecting the motion capture system and the sEMG sensor is that both have high accuracy and are widely used in sports analysis \cite{wang2022technology}. These sensors would be the most suitable, considering this study aims to identify the ground truth differences between standard Bicep Curl and fatigued Bicep Curl. To compare which subset of muscles contributes the most to compensatory movements, we extract muscle synergies from the sEMG signals using NNMF algorithm. 

The contributions of this paper are twofold:
\begin{itemize}
    \item Identification of significant differences in muscle synergy and joint kinematics among weight-free, standard, and fatigued bicep curls.
    \item Proposal of an optimal metric for detecting fatigue-induced compensatory movements in bicep curls, based on a comparison of sEMG data and joint kinematics.
\end{itemize}
The results presented in this paper can potentially guide the optimal development and deployment of wearable sensors for automatic muscle compensation detection in bicep curls in home-based rehabilitation. 
\section{Data Collection}
\label{sec:Data Collection}
\subsection{Participants}
Twelve male participants were recruited in this study. The subjects' age ranged from 18 to 35, their body weight ranged from 65kg to 90kg, and their height ranged from 165cm to 185cm. They had no physical injuries, disabilities, or stroke history. Their self-declared exercise period ranged from one hour to more than seven hours per week. Table \ref{table:Subjects} shows the subjects' demographics. The experiment was approved by the UNSW Human Research Ethics Committee under the reference number HC220573. All subjects provided written consent. Participation in the study was voluntary, and the subjects could terminate their participation at any time before publication.
\begin{table}[h!]
    \centering
    \caption{Subjects Demographics}
    \begin{tabular}{ccccc}
        \hline
        Subject & \makecell{Age \\(year)} & \makecell{Height \\ (cm)} & \makecell{Weight \\(kg)} & \makecell{Exercise duration \\ per week (hour)}\\
        \hline
        
        1 & 18-23 & 175-180 & 60-70 & 3-5\\  
        2 & 18-23 & 175-180 & 70-80 & 5-7\\  
        3 & 24-29 & 175-180 & 70-80 & 3-5\\  
        4 & 30-35 & 175-180 & 80-90 & 1-3\\  
        5 & 24-29 & 175-180 & 60-70 & 5-7\\  
        6 & 18-23 & 170-175 & 70-80 & 7+\\  
        7 & 18-23 & 175-180 & 60-70 & 5-7\\  
        8 & 18-23 & 180-185 & 70-80 & 5-7\\  
        9 & 18-23 & 170-175 & 60-70 & 5-7\\  
        10 & 18-23 & 175-180 & 70-80 & 1-3\\  
        11 & 18-23 & 165-170 & 70-80 & 7+\\  
        12 & 24-29 & 170-175 & 70-80 & 5-7\\ \hline
    \end{tabular}
    \label{table:Subjects}
\end{table}

\subsection{Experimental Setup}
\begin{figure}[th]
    \centering
    \includegraphics[width=0.48\textwidth,height=6cm,keepaspectratio]{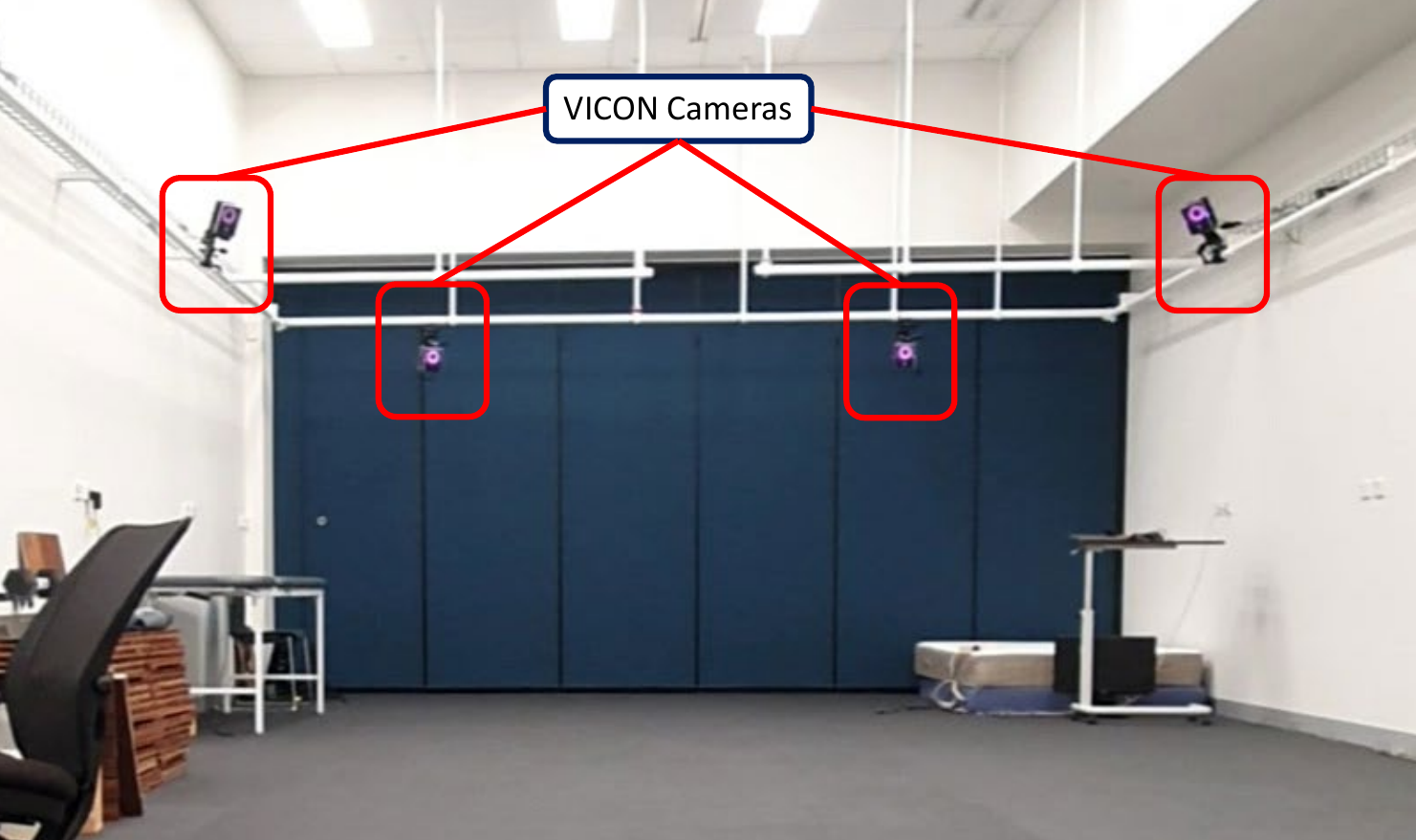}
    \caption{Experimental scene with VICON motion capture cameras.}
    \label{fig:Scene}	
\end{figure}

The experiment was conducted at the Gait Lab of the Falls, Balance, and Injury Research Centre (FBIRC), NeuRA, Australia. The experimental scene is shown in Fig. \ref{fig:Scene}.
	
\begin{figure}[th!]    

    \includegraphics[width=0.45\textwidth]{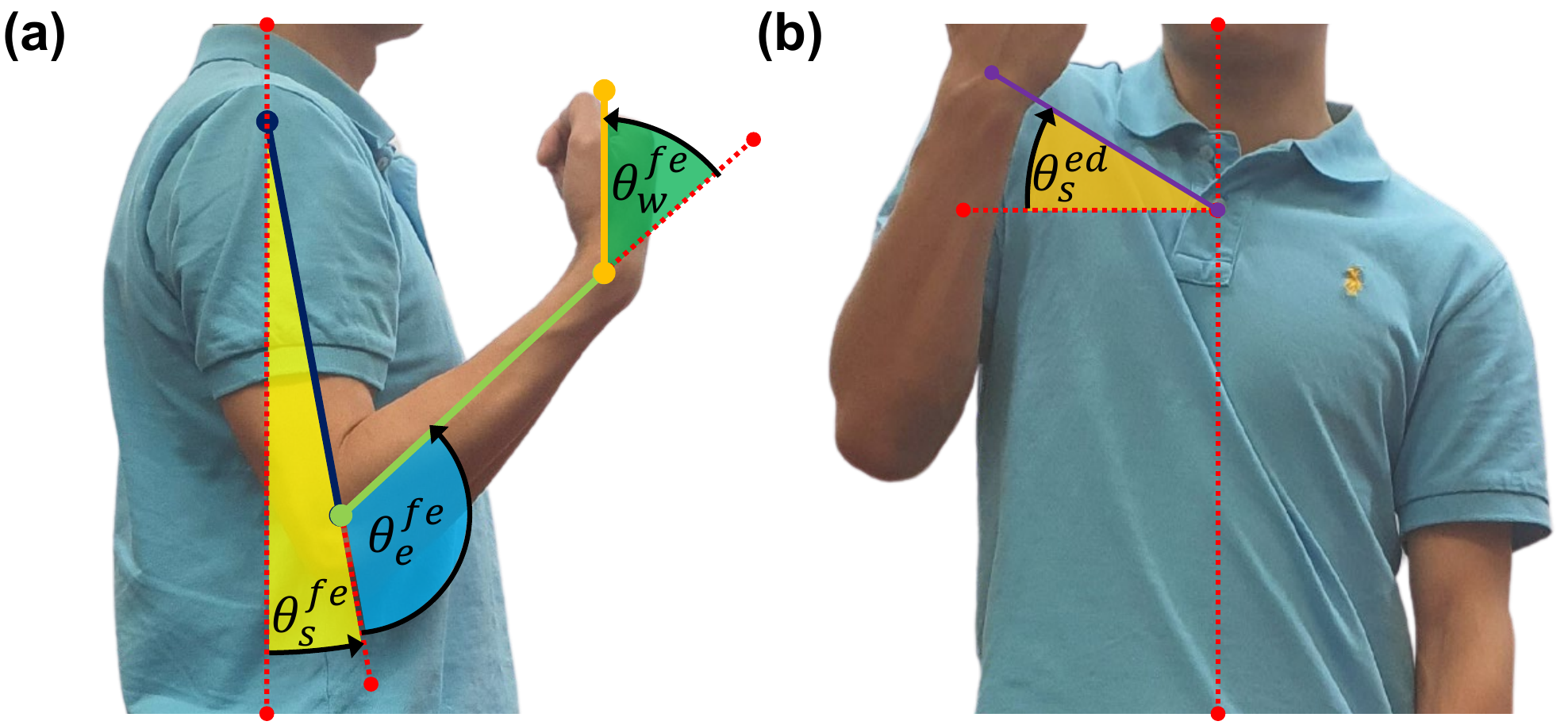}
    \caption{Four joint angles during bicep curl in this study. \textcolor{blue}{(a)} Sagittal view. \textcolor{blue}{(b)} Coronal view. The red dotted lines represent the references; the purple line links the clavicle and the acromioclavicular joint; the blue, green, and yellow lines depict the upper arm limb, the lower arm limb, and the wrist, respectively. The arrows indicate the positive directions.}
    \label{fig:JointDef}
    
\end{figure}

The experiment used eight motion capture cameras (Vantage, VICON Motion Systems, UK) and eight sEMG sensors (Myon 320, zFlo Motion, USA). Thirteen reflective markers were placed on the subjects based on the Upper Limb Model Product Guide \cite{UpperLimbModelVICON} authored by VICON Motion System to capture the subjects' kinematics. The VICON Right Upper Limb model was used to calculate the angles of \textit{shoulder flexion-extension} \(\theta_{s}^{fe}\), \textit{elbow flexion-extension} \(\theta_{e}^{fe}\), \textit{wrist flexion-extension} \(\theta_{w}^{fe}\), and \textit{shoulder elevation-depression} \(\theta_{s}^{ed}\), whose definitions are shown in Fig. \ref{fig:JointDef}. In particular,
\begin{itemize}
    \item \textit{shoulder flexion-extension} \(\theta_{s}^{fe}\) describes the forward and upward movement of the shoulder joint and is defined as the angle between the upper arm and the torso in the sagittal plane (zero degrees represents the state when the shoulder is fully flexed and aligned with the torso); 
    \item \textit{elbow flexion-extension} \(\theta_{e}^{fe}\) denotes the bending movement of the elbow joint and is defined as the angle between the upper arm and the forearm in the sagittal plane (zero degrees represents the state when the elbow is fully flexed);
    \item \textit{wrist flexion-extension} \(\theta_{w}^{fe}\) indicates the bending movement of the wrist joint and is defined as the angle between the forearm and the wrist in the sagittal plane (zero degrees represents the state when the wrist is fully flexed);
    \item \textit{shoulder elevation-depression} \(\theta_{s}^{ed}\) portrays the upward movement of the shoulder joint and is defined as the angle between the line from the clavicle to the shoulder girdle and the line from the clavicle to the C7 vertebra in the coronal plane. 
\end{itemize}
The data were sampled at 100Hz.
	
\begin{figure*}[h!]	
    \includegraphics[width=\textwidth]{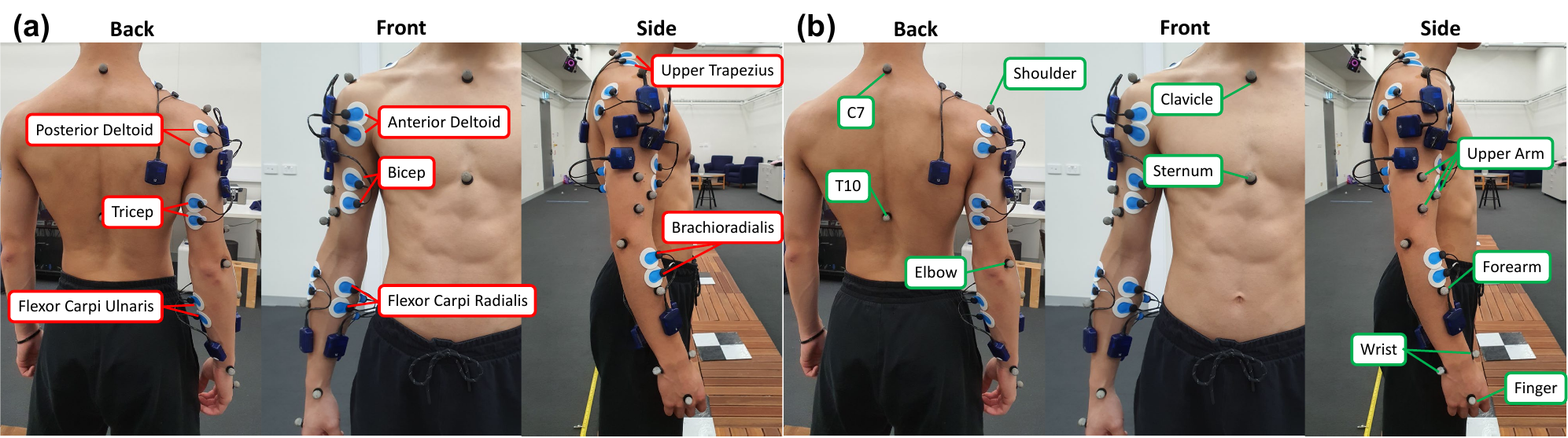}
    \caption{Placement of sEMG sensors and VICON markers on subjects. \textcolor{blue}{(a)} sEMG sensors \textcolor{blue}{(b)} VICON markers}
    \label{fig:image2}
\end{figure*}

The sEMG sensors were placed on eight different muscles of each subject's right arm and shoulder as shown in Fig. \ref{fig:image2}, including \textit{BIC}, \textit{BRA}, \textit{UT}, \textit{TRI}, \textit{FCU}, \textit{FCR}, \textit{AD}, and \textit{PD}. The electrodes were placed based on the SENIAM Guide \cite{Seniam,liu2021muscle}. The data were sampled at 1000Hz.

\subsection{Experimental Protocol}

\subsubsection{Bicep Curl Protocol}
\label{sub:Bic}
The standard bicep curl was selected for this exercise, and the subjects were trained before the experiment to ensure they could perform the same bicep curl variation correctly. The subjects were instructed to supinate their right elbow joint so their palm faced up. Then, the subjects were asked to flex their elbow while keeping the upper arm close to the trunk. When the wrist reached shoulder height, the subjects were instructed to extend the elbow joint back to the starting position. During the study, the subjects were required to do five bicep curls with a 5kg dumbbell following a metronome at 30 beats per minute. In this study, one bicep curl cycle is defined as the time from the subject starts to flex their elbow joint up until the subject's elbow joint returns to the starting position.

\subsubsection{Fatigue Protocol}
Our study uses the Borg CR-10 Scale \cite{liu2021muscle} during the data collection stage to determine fatigue. The subjects were required to self-evaluate themselves during the exercise. The fatigue stage was defined as when the subject's Rated Perceived Exertion (PRE) level was seven (when the subject could not perform a complete cycle or when the subjects orally indicated they could not perform the task). 
From various literature \cite{ABC}, the sEMG signal is proven to be a good metric for reflecting the fatigueness of muscles. The sEMG's amplitude increases and its median frequency decreases when a muscle experiences fatigue. Considering RPE is a subjective, subject-self-rated scale, the maximum root mean square (RMS) amplitude and the median frequency of the sEMG signal were obtained after the data collection to validate the subjects' claim regarding their fatigue state.
	
\subsubsection{Experimental Procedure}
The researcher checked the subjects on the day to ensure they fulfilled all the inclusion criteria. Then, the subject's data, such as weight, height, and exercise habits, were collected, and the researcher explained both the bicep curl protocol and fatigue protocol to the subject. The sEMG sensors and VICON markers were attached at the positions shown in Fig. \ref{fig:image2}. Then, the subject was instructed to perform five repetitions of bicep curls under weight-free conditions, five repetitions under standard conditions with a 5kg dumbbell, and five repetitions under fatigue conditions with a 5kg dumbbell.
	
\section{Methodology}
\subsection{Data Pre-Processing}
The bicep curl measurement was analyzed, and five cycles were taken and averaged to reduce the discrepancies. The sEMG data collected were first filtered using a band-pass filter (30Hz-80Hz) to remove noises \cite{ABC}. The envelope of the signals was then obtained using the `envelope' function from MATLAB. 

\subsubsection{Fatigue Analysis}
To validate the presence of fatigue, the RMS amplitude, \(E_{RMS}\), and the median frequency, \(E_{freq}\), were extracted from the raw sEMG signals \(E_{raw}\) collected at both standard and fatigue conditions. The median frequency of the signal is calculated using the `medfreq' function from MATLAB.
\begin{equation}
	E_{RMS} = \sqrt{\frac{1}{n} \sum_{i} E_{i}^2}
\end{equation}
\begin{equation}
	E_{freq} = \textrm{medfreq}(E_{raw})
\end{equation}

\subsection{Muscle Synergy}
Muscle synergy is a co-activation pattern of muscles recruited by a single neural command signal. It reflects how our muscles work together to make any movement regarding time series and muscle contribution. This study used the default MATLAB `nnmf' function to decompose the filtered sEMG envelopes, \(E\in\mathbb{R}^{m\times t}\), into synergy vectors, \(W\in\mathbb{R}^{m\times n}\), and activation coefficients, \(C\in\mathbb{R}^{n\times t}\), as given by
\begin{equation}
W,C = \textrm{nnmf}(E), \quad \min_{W,C} |W\times C - E| 
\end{equation}
where \(m\), \(n\), and \(t\) represent the number of muscles observed, the number of synergies, and the number of data points, respectively. For more details of this method, readers are referred to \cite{turpin2021improve}.

\subsubsection{Relative Muscle Synergy}
After extracting the muscle synergy for all subjects, the data were normalized using L2 Normalization \cite{turpin2021improve} to values between [0 1] to allow inter-subject comparison. The inverse of the normalized coefficient, \(D\), was then integrated into the activation. The mathematical model of L2 Normalization and the computation of relative muscle synergy \(w\) and activation pattern \(c\) are given by
\begin{equation}
    w = W \times D^{-1}, \quad c = D \times C
\end{equation}
where
\begin{equation}
    D = \sqrt{\sum_{i=1}^{n}\norm{W_{i}}}
\end{equation}
	
\subsubsection{Variance Accounted For (VAF)}
The number of muscle synergies was determined by the VAF value between the product of the weight synergy coefficient, the activation coefficient, and the sEMG envelope. The number of synergies was defined as the minimum number of synergies underlying VAF $\ge$ 0.9 while the growth rate was less than 3\%. The mathematical model of VAF is shown below.
\begin{equation}
	VAF = 1 - \frac{(E-w \times c)^2}{E^2}
	\label{eq:VAF}
\end{equation}
	
\subsubsection{Cosine Similarity}
Cosine similarity, measuring the cosine angle between two vectors, was used to compare and match synergies. The equation is shown below.
\begin{equation}
    \textrm{cos}(V_1,V_2) = \frac{V_1\cdot V_2}{\left\| V_{1}\right\| \left\| V_{2}\right\|}  
\end{equation}

\subsubsection{Single Muscle Fatigue Identification}
After computing the muscle synergy of the Bicep Curl under different conditions, we can identify the changes in the relative contributions among the selected muscles. The changes can be interpreted as the changes in the muscle recruitment strategy from the central nervous system to maintain performance to conduct the exercise. Hence, we selected the subset of muscles with the most significant changes in the muscle synergy model and tabulated their increment in the RMS amplitude.

\subsection{Kinematics}
The VICON software outputs various joint kinematics based on the coordinates of VICON markers and the VICON Upper Limb Model, including shoulder flexion-extension, abduction-adduction, elevation-depression, internal-external rotation, elbow flexion-extension, and wrist flexion-extension. Then, these joint angle trajectories were filtered using the move mean algorithm at a window size of 1\% of the overall data length. 

\subsubsection{Time Scale Normalization}
The joint kinematics observed during the study, \(J\in\mathbb{R}^{j\times d}\), was segmented and normalized using spline interpolation method, where \(j\) and \(d\) represent the number of joint kinematics observed and the data length, respectively. The joint kinematics \(J\) was normalized into \(\ell\) number of data points based on the definition of the bicep curl described in the section~\ref{sub:Bic} to allow comparison. The bicep curl was divided into the lifting-up phase and the lowering-down phase. The lifting-up phase ranged from 0\% to 50\%, and the lowering-down phase ranged from 50\% to 100\%. The normalization was conducted using the `interp1' function from MATLAB.
\begin{equation}
	\left\| J\right\|_t := \textrm{interp1}(J, \ell)
\end{equation}

\subsubsection{Similarity \& Discrepancy Among Subjects}
After normalizing the time scale of the bicep curl conducted by all subjects, the averaged normalized joint kinematics \(\bar{J}\in\mathbb{R}^{j\times \ell}\) is obtained as
\begin{equation}
	\bar{J} = \frac{1}{s} \sum_{i=1}^{s} \left\| J_{i}\right\|_t .
\end{equation}
where \(s\) represents the total number of subjects.

Then, the average difference for all data points, $\mathcal{J}\in\mathbb{R}^{j\times \ell}$ between the averaged joint kinematics $\bar{J}$ and each subject's individual joint kinematics $\left\| J\right\|_t$ is given by
\begin{equation}
    \mathcal{J} = \omega(\bar{J},\left\| J\right\|)
\end{equation}
where $\omega(\cdot,\cdot)$ is an auxiliary function defined as,
\begin{equation}
    \omega(x,y):= \frac{1}{n}\sum_{i=1}^{n} (x_{i}-y_{i}), \quad \forall \, x,y\in\mathbb{R}^n.
\end{equation}

Then, the similarity, \(\overline{\mathcal{J}}\) and discrepancy, \(\sigma_{\mathcal{J}}\) of the joint kinematics relative to their averaged joint kinematics were assessed by calculating the mean and standard deviation of \(\mathcal{J}\), as follow:
\begin{equation}
 	\overline{\mathcal{J}} = \frac{1}{s}\sum_{i=1}^{s}\mathcal{J}_{i}
\end{equation}
\begin{equation}
    \sigma_{\mathcal{J}} = \sqrt{\frac{1}{s-1} \sum_{i=1}^s (\mathcal{J}_{i} - \overline{\mathcal{J}})^2}
\end{equation}
\subsection{Statistical Analysis}
MATLAB was used for statistical analysis with the p-value set at 5\%. The normality of the data was tested using the Shapiro-Wilk test, and the equality of variance was tested using Levene's test. Pair t-test is the most suitable test if both conditions apply. Otherwise, Wilcoxon signed rank test should be used for this purpose. Due to the low sample size of this study (12), it is lower than the general guidelines of the minimum sample size (30) to be considered sufficient for the Central Limit Theorem to hold. Hence, the Wilcoxon signed rank test was used, as the data compared were paired and did not require the data to be normally distributed.

\section{Result}
This research presents the bicep curl's muscle synergy and kinematic movements under weight-free, standard, and fatigue conditions. We also provide the joint kinematics under these modes. Furthermore, we discuss the general differences between the bicep curl of these states and the muscles required to observe and capture the change between these states.
	
\subsection{Median Frequency \& RMS Analysis of EMG}
\begin{figure*}[h!]
	\centering
	\includegraphics[width=1\textwidth,keepaspectratio]{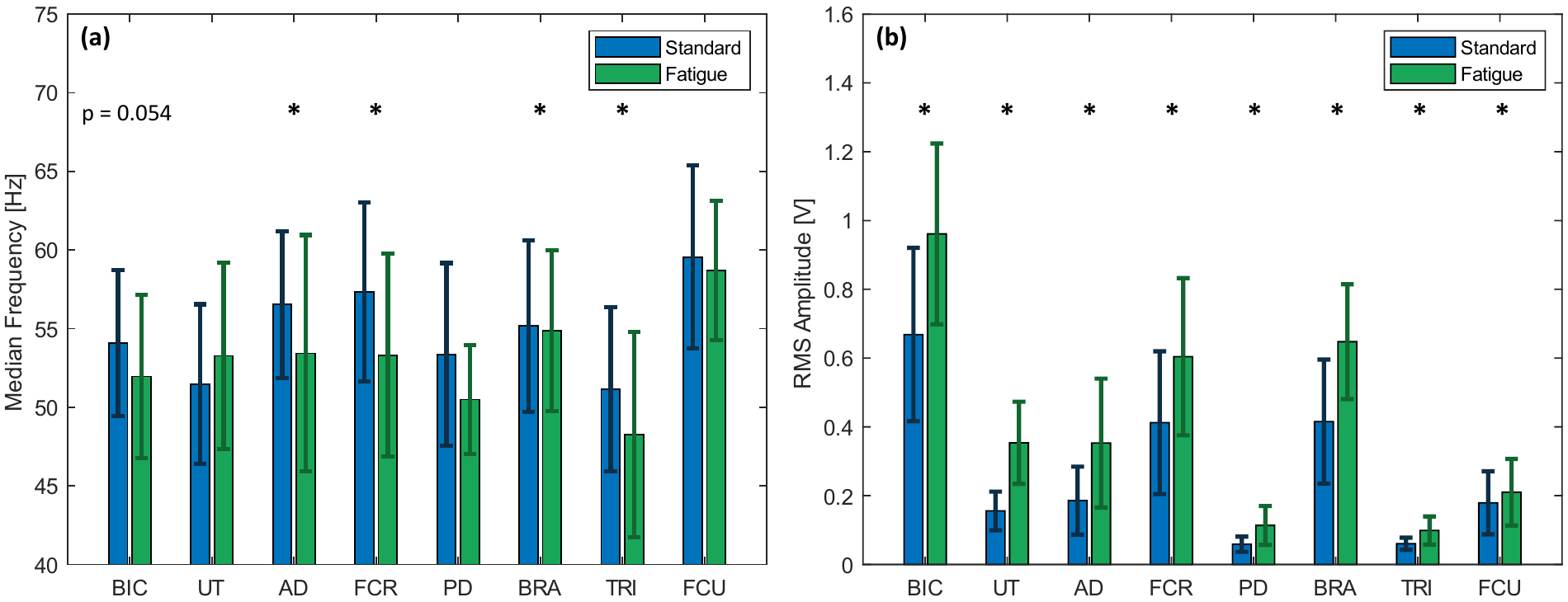}
	\caption{The fatigue metrics. \textcolor{blue}{(a)} The averaged median frequency of the subjects. \textcolor{blue}{(b)} The averaged RMS amplitude of the subjects. The blue bars and error bars show the average median frequency and RMS amplitude of the sEMG collected under standard conditions. In contrast, the green bars and error bars represent the data under fatigued conditions. }
	\label{fig:Fatigue}
		
\end{figure*}

This section outlines the changes in the median frequency and RMS amplitude of the signal of all muscles between standard and fatigued bicep curls, as shown in Fig. \ref{fig:Fatigue}. The result shows a significant reduction in the median frequency of the \textit{AD}, \textit{FCU}, \textit{PD}, and \textit{TRI} muscles. The median frequency of all muscles experienced a reduction of 0.50\% to 7.01\% in the fatigue conditions compared to the standard conditions. The median frequency of half of the muscles under two conditions revealed a statistically significant difference (p-value $<$ 0.05) using the Wilcoxon signed rank test, supporting that there was a measurable difference between the median frequency of the muscles under standard and fatigue states under all subjects. The \textit{BIC} muscle recorded a p-value of 0.054, which is moderately close to the threshold. Notably, the antagonist muscle \textit{TRI} and synergist muscle \textit{BRA} have reported fatigue which fulfills Boyas et al. \cite{boyas2009changes} and Wang et al. \cite{wang2022antagonist} findings on the changes of synergist muscles and antagonist muscle in the frequency domain during repetitive exercise-induced fatigue. Hence, we can conclude that the subjects have achieved fatigue.

The results show that the RMS amplitude of all muscles' sEMG signals had a significant increment from 17.2\% to 127.2\% comparing the standard and the fatigue conditions. The effect was tested using the Wilcoxon signed rank test, revealing that the increment for all muscles in the RMS amplitude is statistically significant (p-value $<$ 0.05). Theoretically, the RMS of sEMG signal would be affected by fatigue and the force exerted by the muscle. Considering the exercise intensity, ROM, and load remain constant, we may conclude that the changes in RMS include fatigue factors and changes in muscle recruitment strategy to maintain the performance of the exercise.

\subsection{Minimum Muscle Synergy}
\begin{figure}[h!]
	\centering
	\includegraphics[width=0.45\textwidth]{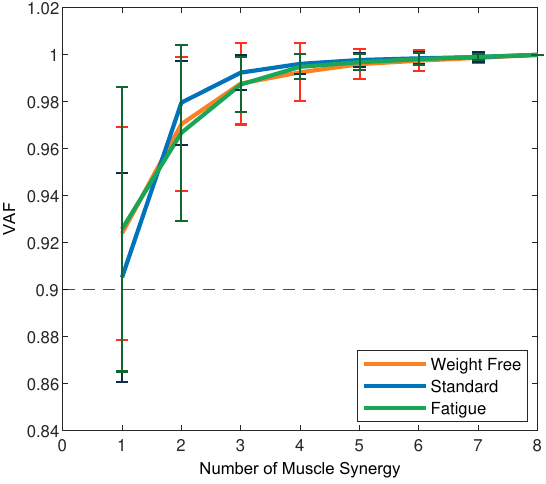} 
	\caption{Averaged VAF values of different numbers of synergy for different stages. The orange graph represents the averaged VAF values for weight-free conditions; the blue graph represents the averaged VAF values for standard conditions, and the green graph represents the averaged VAF values for fatigue conditions.}  
	\label{fig:VAF}
		
\end{figure}

The overall VAF value with different numbers of muscle synergies for all conditions is shown in Fig. \ref{fig:VAF}. It was calculated based on equation \ref{eq:VAF}. The results show that the VAF value hits 90\%, and the growth rate is not over 3\% when two is selected as the minimum number of muscle synergies for all conditions. It shows that the changes in the conditions do not change the complexity of the muscle coordination. Hence, we extracted two muscle synergies across all subjects and conditions.
	
\subsection{Muscle Synergy Analysis}
\begin{figure*}[t!]
	\centering
	\includegraphics[width=\textwidth]{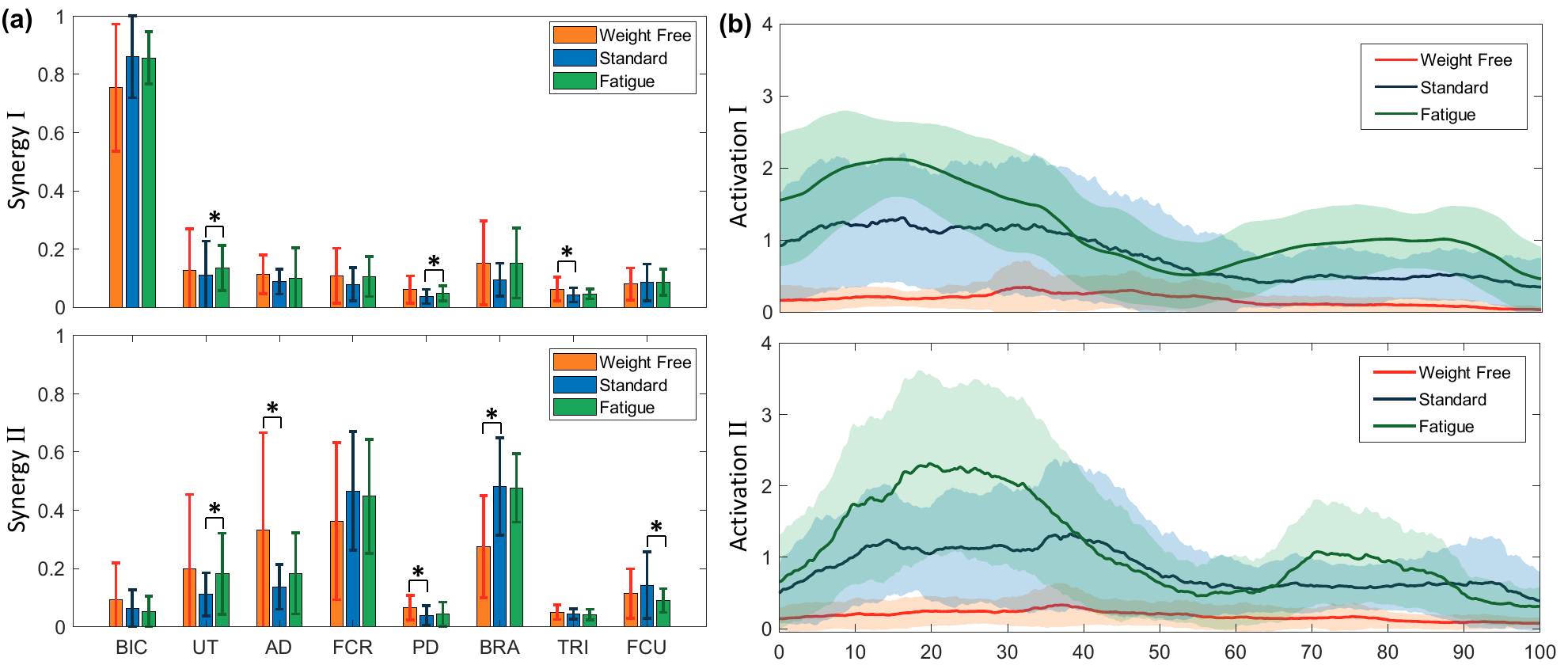} 
	\caption{Muscle Synergy and Activation Profiles between stages. \textcolor{blue}{(a)} Muscle Synergy Comparison between stages. The orange bar represents the averaged synergy under the weight-free condition; the blue bar represents the averaged synergy under the standard condition; the green bar represents the averaged synergy under the fatigue condition. The error bar of each bar chart represents the standard deviation of the data. The \(*\) sign on top of the bar represents the statistical difference between the datasets. \textcolor{blue}{(b)} Activation Profiles between stages. The orange line represents the average activation profile under the weight-free condition; the blue line represents the average activation profile under the weighted condition; the green line represents the average activation profile under the fatigue condition. The shaded region represents the standard error.}
    \label{fig:Syn}
\end{figure*}

The muscle synergy patterns across all subjects were extracted and shown in Fig. \ref{fig:Syn}. The NNMF algorithm outputted the synergies and activation patterns, which were not arranged in any order. Using cosine similarity, we differentiated and grouped the muscle synergy in a specific order. We define Synergy I as the synergy with the highest relative contribution of the \textit{BIC} muscle. The remaining is defined as Synergy II.
Synergy I for all stages consisted of a high weighting of BIC muscle contribution ranging from 75\% to 86\% of the total weighting, representing the predominant movement of this study, elbow flexion. 
	
\subsubsection{Weight-Free Bicep Curl}
In this group, the average VAF was 94.5\% with a standard deviation of 2.38\%. Synergy II consisted of relatively balanced contributions among shoulder and forearm muscles. \textit{UT}, \textit{AD}, \textit{FCR}, and \textit{BRA} muscles have similar relative contributions ranging from 19.9\% to 36.2\%.
	
\subsubsection{Standard Bicep Curl}
Under the standard condition,  we noticed significant changes in several muscle contributions under Synergy II compared to weight-free bicep curl. We observed a decrease in the relative contribution of the shoulder muscles, such as \textit{UT} and \textit{AD}, from 19.9\% and 33.3\% to 11.2\% and 13.8\%. We noticed an increment in the relative contribution of the forearm muscles, such as \textit{FCR} and \textit{BRA}, from 36.2\% and 27.5\% to 46.6\% and 48.1\%. However, only the changes in the \textit{AD}, \textit{PD}, and \textit{BRA} muscles are statistically significant based on the Wilcoxon signed rank test. In addition, there were no visible changes in the relative contribution of muscles in Synergy I. Notably, the \textit{TRI} muscle had a statistically significant decrement from 6.25\% to 4.56\% in its relative contribution.
	
\subsubsection{Fatigue Bicep Curl}
Under the fatigue condition, we noticed significant changes in several muscle contributions under Synergy II compared to standard bicep curls. We observed an increment in the relative contribution of the shoulder muscles, such as \textit{UT} and \textit{AD}, from 11.2\% and 13.8\% to 18.2\% and 18.3\%. We also noticed a minor decrement in the relative contribution of the forearm muscles, such as \textit{FCR} and \textit{FCU}, from 46.6\% and 14.3\% to 44.7\% and 9.06\%. Using the Wilcoxon signed rank test, only the changes in the relative contribution of the \textit{UT} and \textit{FCU} muscles are statistically significant. Similar to the standard condition, there were no visible changes in the relative contribution of muscles in Synergy I. Notably, the \textit{PD} and \textit{UT} muscles under this condition have significant differences compared to the standard bicep curl condition.

\subsubsection{Activation}
The result in this section demonstrated the averaged activation patterns among subjects. 
For Synergy I, the RMS amplitude of the activation under the weight-free condition was around 0.19, and it did not change throughout the bicep curl cycle. Under the standard condition, the RMS amplitude increased to around 0.86, and we can see a steep decrement at around 50\% of the bicep curl and then stabilize after. Lastly, the RMS amplitude of the activation under fatigue conditions increased to approximately 1.3. Under this condition, the shape appeared to be a sin wave, having an early start, achieving the first peak at around 10\% of the bicep curl, reaching the first through at around 50\% of the bicep curl, and then forming a second curve with about half of the amplitude of the first curve.
	
For Synergy II, we observed a similar activation under weight-free conditions as Synergy I. Under the standard condition, the RMS amplitude increased to around 0.88, and it occurred to have an increasing trend from the start to around 20\% of the bicep curl, remained constant till approximately 50\% of the bicep curl, and then reduced after. Lastly, the RMS amplitude of the activation under fatigue conditions increased to about 1.3. The shape appeared to be a sinusoidal wave, achieving the first peak at around 30\% of the bicep curl, reaching the first through at around 60\% of the bicep curl, and then forming a second peak, with half of the amplitude of the first peak. 
	
\subsection{Kinematics}
\begin{figure*}[h!]
    \centering
    \includegraphics[width=\textwidth]{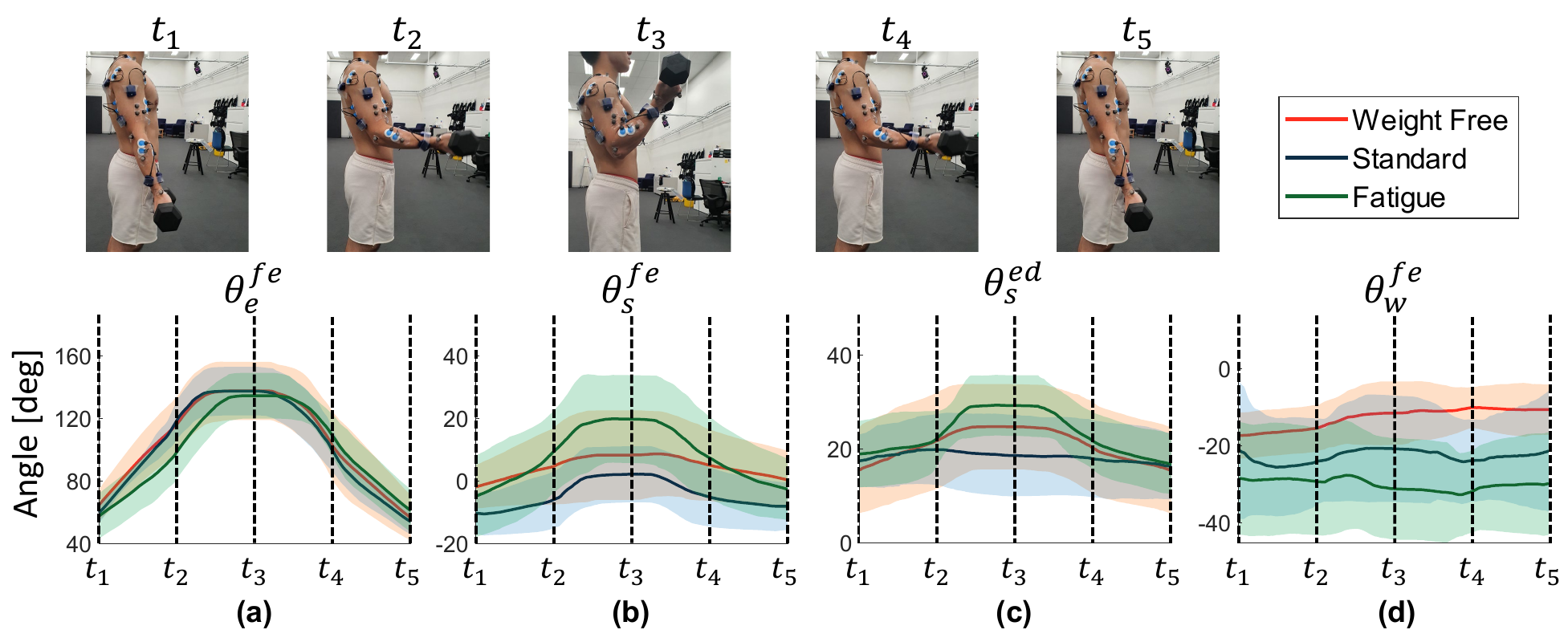} 
    \caption{Joint Kinematics of the Bicep Curl. The Bicep Curl exercise is segmented into five stages. The images show the stage when the subject conducts Bicep Curl at each stage. The orange line represents the averaged joint kinematics under weight-free conditions; the blue line represents the averaged joint kinematics under standard conditions; the green line represents the averaged joint kinematics under fatigue conditions. Their respective shaded area represents the standard error of the trajectory.
    \textcolor{blue}{(a)} \textit{Elbow Flexion - Extension}
    \textcolor{blue}{(b)} \textit{Shoulder Flexion - Extension}
    \textcolor{blue}{(c)} \textit{Shoulder Elevation - Depression}
    \textcolor{blue}{(d)} \textit{Wrist Flexion - Extension}}
    \label{fig:Kin}
\end{figure*}

In this study, joint kinematics such as \textit{elbow flexion-extension}, \textit{shoulder flexion-extension}, \textit{shoulder elevation-depression}, and \textit{wrist flexion-extension} were collected. At the same time, the subjects conducted bicep curl exercises under various conditions. The result is shown in Fig. \ref{fig:Kin}, and the difference between each subject's kinematics and the average kinematics is tabulated in Table. \ref{fig:JointAcc}.
\subsubsection{Elbow Flexion - Extension}
We noticed a similar range of motion from 50 to 150 degrees for \textit{elbow flexion-extension}. The trajectory for both weight-free and weight had no visible difference. For the fatigue conditions, we noticed a right shift during the lifting-up stage from reaching the peak at around 30\% to 40\% of the bicep curl. The accuracy and similarity among subjects were the lowest for the weight-free condition, followed by the fatigue condition, and then the standard condition.
\subsubsection{Shoulder Flexion - Extension}
Comparing all three stages, the shoulder flexed in a similar trajectory. 
Under the weight-free condition, the shoulder remained stable, and barely any flexion occurred. Under the standard condition, we noticed the starting position was lowered at around -10 degrees, showing that the subjects flexed slightly behind the center of the trunk due to the presence of the dumbbell. The shoulder slightly flexed for approximately 10 degrees at around 30\% and returned at around 60\% of the bicep curl cycle. Under the fatigue condition, we noticed the shoulder flexed more vigorously compared to the standard condition for about 20 degrees at around 20\% and returned at around 80\% of the bicep curl cycle. 
The accuracy and similarity among subjects were the lowest under the fatigue condition, followed by the weight-free condition and then the standard condition.
\subsubsection{Shoulder Elevation - Depression}
Under the weight-free condition, the shoulder started with an angle of around 15 degrees, elevated at around 30\% of the bicep curl, reached 25 degrees at around 40\% of the bicep curl, and returned to around 15 degrees at around 70\% of the bicep curl. Under the standard conditions, the shoulder remained stable throughout the cycle, starting at an angle around 16 degrees, peaking at 20 degrees at about 25\% of the bicep curl. Under the fatigue condition, we noticed similar patterns as the weight-free condition. The difference was that the shoulder elevated more vigorously to 30 degrees at around 20\% of the bicep curl cycle and returned to the initial level at around 70\% of the cycle. In terms of accuracy and similarity among subjects, it was the lowest under the weight-free condition, followed by the standard condition, and then the fatigue condition.

\subsubsection{Wrist Flexion - Extension}
Unlike the shoulder joints, the wrist flexion angle for all three conditions remained constant throughout all conditions. The only difference was that the initial angle for all three stages decreased across each condition. The initial wrist flexion-extension angle for weight-free condition, standard condition, and fatigue condition were approximately -17 degrees, -21 degrees, and -29 degrees, respectively. Regarding accuracy and similarity among subjects, it was the highest under the weight-free condition, followed by the standard and the fatigue conditions.

\begin{table}[h!]
		\caption{Similarity and Discrepancy between Individual Subjects and Averaged Joint Trajectory.  }
            \centering
		\begin{tabular}{ccccccc}
                \hline 
                
\multicolumn{1}{c}{} & \multicolumn{2}{c}{Weight Free}  & \multicolumn{2}{c}{Standard}  & \multicolumn{2}{c}{Fatigue}                                           \\ \cline{2-7} 
\multicolumn{1}{c}{\multirow{-2}{*}{\begin{tabular}[c]{@{}c@{}}Joint \\ Trajectory\end{tabular}}} & \multicolumn{1}{c}{ \makecell{\(\overline{\mathcal{J}}\)  \\(deg)}} & \multicolumn{1}{c}{\makecell{\(\sigma_{\mathcal{J}}\)  \\(deg)}} & \multicolumn{1}{c}{ \makecell{\(\overline{\mathcal{J}}\)  \\(deg)}} & \multicolumn{1}{c}{\makecell{\(\sigma_{\mathcal{J}}\)  \\(deg)}} & \multicolumn{1}{c}{ \makecell{\(\overline{\mathcal{J}}\)  \\(deg)}} & \multicolumn{1}{c}{\makecell{\(\sigma_{\mathcal{J}}\)  \\(deg)}} \\ \hline
\(\theta_{e}^{fe}\)&14.2&18.2&9.39&11.8&11.8&14.8 \\
\(\theta_{s}^{fe}\)&8.60&11.7&6.38&8.71&10.7&13.4 \\
\(\theta_{s}^{ed}\)&7.61&9.51&6.47&7.85&5.32&6.45 \\
\(\theta_{w}^{fe}\)&5.17&6.67&10.5&12.8&10.2&13.2 \\ \hline

		\end{tabular}
     \label{fig:JointAcc}
\end{table}
\section{Discussion}
This study is the first to examine the differences in the coordination of the shoulder and limb muscles when the subjects are conducting bicep curls under weight-free, standard, and fatigue conditions. We observed a positive correlation in the RMS and median frequency with the subjects' RPE scale from the result. We noticed from the median frequency analysis that the synergist muscles, some fixator muscles, and the antagonist muscles of the bicep curl exercise experienced fatigue.

\subsection{Muscle Synergy}
The number of synergies found for all conditions was two, showing no significant changes in muscle coordination complexity.
From the result section, we observed no significant difference in Synergy I across different conditions; hence, the discussion will be focused on Synergy II. 
In our study, when the subjects conducted bicep curl exercises, we noticed a shift in the muscle contribution from shoulder to forearm muscles, which was statistically significant. The presence of the weight decreased the stability of the wrist, and it required more force to adapt to the increasing effort.
When the subjects conducted bicep curl exercises under the fatigue conditions, we noticed a shift in the muscle contribution from the forearm muscles to the shoulder muscle. In contrast to Liu et al. \cite{liu2021muscle} result, which demonstrated the increasing coherence of the \textit{FCU}, \textit{FCR}, and \textit{BRA} muscles over fatigue, we found that the load shifted towards the shoulder muscles over the forearm muscles. However, their research did not include shoulder muscles, which may have led to that conclusion.

\subsection{Activation Patterns}
Compared to synergy patterns, the activation patterns provided more insightful information. Across all conditions, the amplitude of the activation profiles increased. This alteration was caused by the increasing effort to conduct bicep curl exercises over conditions. 
For Synergy I, the muscles were activated at the beginning of the bicep curl, indicating the concentric phase of the bicep curl. At the isometric stage, around 50\% of the bicep curl, the activation showed a sharp decrease and then remained constant at the eccentric phase. For Synergy II, the muscles gradually activated during the concentric phase due to the increasing effort to stabilize the grip while the dumbbell was away from the trunk. At the isometric stage, around 50\% of the bicep curl, the activation remained constant at the eccentric phase. Under the fatigue condition, Synergy I was highly activated initially to compensate for the reduced force-generating capacity. Once the movement was initiated, the activation decreased, reaching the first through at the isometric phase. During the eccentric phase, the activation gradually increases to control the descent of the dumbbell. In Synergy II, the activation gradually increased and reached the peak at around 25\% and 75\% of the bicep curl, which is the furthest distance between the dumbbell and the trunk during the concentric and eccentric phases to provide extra grip.

\subsection{Kinematics}
From the result, the kinematic measures for bicep curl (elbow flexion) did not significantly change the range of motion. However, the average angle difference for weight-free conditions was the highest, which may be caused by the difficulty in maintaining a consistent cadence while conducting bicep curls. In addition, we noticed visible changes under different conditions, notably the shoulder joint. The result demonstrated significant changes in \textit{shoulder flexion-extension} and \textit{shoulder elevation-depression} movement. The central nervous system regulates the shoulder to aid the bicep curl action. Interestingly, the angle difference and the discrepancy for \textit{shoulder elevation-depression} movement under the fatigue condition was the lowest compared to other conditions, showing that the subjects tended to compensate for fatigue by elevating the shoulder. On the other hand, the angle difference and the discrepancy for \textit{shoulder flexion-extension} under the fatigue condition was the highest compared to other conditions, showing that the introduction of excessive shoulder flexion under fatigue was unstable and inconsistent. Lastly, the result demonstrated that the wrist became more unstable over various conditions. 

\subsection{Insights for Sensor Development}
From the result, we noticed a statistically significant increment in relative contribution among shoulder muscles, especially \textit{UT}, between standard and fatigue conditions. On the other hand, we noticed a statistically significant decrement among forearm muscles, especially \textit{FCU}. Bringing this information to the RMS amplitude plot and the median frequency plot in Fig. \ref{fig:Fatigue}, we can conclude that when the subject is experiencing fatigue, the RMS amplitude of the sEMG signal from \textit{UT} and \textit{FCU} will experience an increment of 127\% and 17.2\% compared to standard form. However, the changes in the median frequency of both muscles are not statistically significant and hence it does not conclude that both muscles were experiencing fatigue. The changes in the RMS should mainly consist of the change in the contribution of muscle force. However, we do not recommend selecting \textit{FCU} as an optimal metric as the changes in RMS are insignificant.

Similarly, we observed profound movement in the shoulder. The range of motion of shoulder elevation-depression has increased approximately 6 degrees from [16,20] degrees to [20,30] degrees, while the range of motion of shoulder flexion-extension has increased approximately 10 degrees from [-10,0] degrees to [0,20] degrees. Both data are significant for capturing the transition from standard to fatigue. In short, our findings found that when fatigue occurs:
\begin{itemize}
    \item The relative contribution of \textit{UT} muscle increased by around 62.5\%, and the RMS amplitude of the sEMG signal for \textit{UT} muscle increased by 127\%.
    \item The shoulder joint has more vigorous movement, including increased amplitude in both \textit{shoulder elevation-depression} motion and \textit{shoulder flexion-extension} motion.

\end{itemize}

In terms of biomechanics, the upper trapezius muscle is the primary mover for shoulder elevation movement. We may combine these findings and conclude that the central nervous system recruits the upper trapezius muscle to execute the Bicep Curl exercise, leading to excessive shoulder elevation-depression movement. 
From the findings listed above, we may provide a suggestion for detecting fatigue during Bicep Curl exercise using different metrics. Having the prior knowledge that the subject is conducting bicep curl, without changing load and maintaining constant ROM, future studies can detect fatigue via:
\begin{itemize}
   \item Collecting the sEMG signal or monitoring the muscle force of the \textit{UT} muscle. It is expected to have a significant increment when fatigue occurs. Or
   \item Collecting the ROM of the shoulder joint kinematics. It is expected to have a significant increment when fatigue occurs.
\end{itemize}                    

\subsection{Limitation}

This study has focused on muscles and joints of the upper limb directly relating to Bicep Curls. Future work that directly expands from this study includes utilizing different sensors to develop a wearable sensor to identify the patient's state. Relevant sensors include inertial measurement units, cameras, sEMG sensors, mechanomyography (MMG) sensors, and bending sensors \cite{guo2021human}.
During the investigation, we instructed the subjects to stand still and limit the movement to the upper limb, trying to isolate the contribution of other muscles. However, other muscles, such as core trunk muscles, may also contribute to the compensatory movements in bicep curls. It might be worth further investigating the role of those muscles in the compensation of bicep curls. Furthermore, future research can utilize biomechanical modeling software such as OpenSim and Anybody to explore the muscle contributions of Bicep Curl exercise under different conditions such as speed, grip, and weight. In addition, the EMG sensor is highly sensitive; its placement on the muscle is exceptionally crucial. Considering our suggestion for wearable sensor development, which includes examining the UT muscle using an sEMG sensor, the patients might likely misplace it due to not being professionally trained, and this study did not consider the effect of sensor misplacement on the result. Future studies can design wearable sensors to reduce or investigate the effects of misplacement.

\section{Conclusion}
This study examined the muscle synergies and the joint kinematics during weight-free, standard, and fatigue conditions. Compared to the standard stage, our findings indicate a shift in relative contribution from the forearm muscles to the shoulder muscles, and the activation amplitude for both synergies is significantly increased. In addition, more pronounced movement is captured on the shoulder joint during fatigue. This work offers valuable insights into identifying fatigue-induced muscle compensation in sports science and rehabilitation and establishes numerical benchmarks for sensor development that capture the fatigue stage. Based on the findings of this work, the team is investigating new strain-based wearable sensors to be deployed around the shoulder for automatically detecting fatigue-induced compensatory movements in bicep curls.

\section{Acknowledgement}
We express our gratitude to Dr Nan Hu (UNSW School of Clinical Medicine) for advice on statistical analysis and A/Prof Mohit Shivdasani and Dr Luca Modenese (UNSW Graduate School of Biomedical Engineering) for advice on sEMG processing. We also thank the UNSW School of Mechanical and Manufacturing Engineering and NeuRA Falls, Balance, and Injury Research Centre for the resources provided to complete the study. 
\bibliographystyle{IEEEtran}
\bibliography{ref.bib}

\begin{thebibliography}{10}
\providecommand{\url}[1]{#1}
\csname url@samestyle\endcsname
\providecommand{\newblock}{\relax}
\providecommand{\bibinfo}[2]{#2}
\providecommand{\BIBentrySTDinterwordspacing}{\spaceskip=0pt\relax}
\providecommand{\BIBentryALTinterwordstretchfactor}{4}
\providecommand{\BIBentryALTinterwordspacing}{\spaceskip=\fontdimen2\font plus
\BIBentryALTinterwordstretchfactor\fontdimen3\font minus
  \fontdimen4\font\relax}
\providecommand{\BIBforeignlanguage}[2]{{%
\expandafter\ifx\csname l@#1\endcsname\relax
\typeout{** WARNING: IEEEtran.bst: No hyphenation pattern has been}%
\typeout{** loaded for the language `#1'. Using the pattern for}%
\typeout{** the default language instead.}%
\else
\language=\csname l@#1\endcsname
\fi
#2}}
\providecommand{\BIBdecl}{\relax}
\BIBdecl

\bibitem{churgay2009diagnosis}
C.~A. Churgay, ``Diagnosis and treatment of biceps tendinitis and tendinosis,''
  \emph{American family physician}, vol.~80, no.~5, pp. 470--476, 2009.

\bibitem{physiopedia2022rehab}
\BIBentryALTinterwordspacing
Physiopedia, ``Rehabilitation in sport --- physiopedia{,},'' 2022, [Online;
  accessed 17-January-2024]. [Online]. Available:
  \url{https://www.physio-pedia.com/index.php?title=Rehabilitation_in_Sport&oldid=323414}
\BIBentrySTDinterwordspacing

\bibitem{levin2009motor}
M.~F. Levin, J.~A. Kleim, and S.~L. Wolf, ``What do motor “recovery” and
  “compensation” mean in patients following stroke?''
  \emph{Neurorehabilitation and neural repair}, vol.~23, no.~4, pp. 313--319,
  2009.

\bibitem{van2004continuity}
C.~Van~Walraven, M.~Mamdani, J.~Fang, and P.~C. Austin, ``Continuity of care
  and patient outcomes after hospital discharge,'' \emph{Journal of general
  internal medicine}, vol.~19, no.~6, pp. 624--631, 2004.

\bibitem{moksnes2023factors}
H.~{\O}. Moksnes, C.~Sch{\"a}fer, M.~S. Rasmussen, H.~L. S{\o}berg,
  O.~R{\o}ise, A.~Anke, C.~R{\o}e, P.~A. N{\ae}ss, C.~Gaarder, E.~Helseth
  \emph{et~al.}, ``Factors associated with discharge destination from acute
  care after moderate-to-severe traumatic injuries in norway: a prospective
  population-based study,'' \emph{Injury epidemiology}, vol.~10, no.~1, p.~20,
  2023.

\bibitem{lin2022vr}
M.~Lin, J.~Huang, J.~Fu, Y.~Sun, and Q.~Fang, ``A vr-based motor imagery
  training system with emg-based real-time feedback for post-stroke
  rehabilitation,'' \emph{IEEE Transactions on Neural Systems and
  Rehabilitation Engineering}, vol.~31, pp. 1--10, 2022.

\bibitem{arntz2023technologies}
A.~Arntz, F.~Weber, M.~Handgraaf, K.~L{\"a}ll{\"a}, K.~Korniloff, K.-P.
  Murtonen, J.~Chichaeva, A.~Kidritsch, M.~Heller, E.~Sakellari \emph{et~al.},
  ``Technologies in home-based digital rehabilitation: Scoping review,''
  \emph{JMIR rehabilitation and assistive technologies}, vol.~10, no.~1, p.
  e43615, 2023.

\bibitem{wang2022technology}
X.~Wang, Y.~Fu, B.~Ye, J.~Babineau, Y.~Ding, and A.~Mihailidis,
  ``Technology-based compensation assessment and detection of upper extremity
  activities of stroke survivors: Systematic review,'' \emph{Journal of Medical
  Internet Research}, vol.~24, no.~6, p. e34307, 2022.

\bibitem{nicholson2020multi}
C.~Nicholson-Smith, V.~Mehrabi, S.~F. Atashzar, and R.~V. Patel, ``A
  multi-functional lower-and upper-limb stroke rehabilitation robot,''
  \emph{IEEE Transactions on Medical Robotics and Bionics}, vol.~2, no.~4, pp.
  549--552, 2020.

\bibitem{liu2022home}
Y.~Liu, S.~Guo, Z.~Yang, H.~Hirata, and T.~Tamiya, ``A home-based
  tele-rehabilitation system with enhanced therapist-patient remote
  interaction: A feasibility study,'' \emph{IEEE Journal of Biomedical and
  Health Informatics}, vol.~26, no.~8, pp. 4176--4186, 2022.

\bibitem{darmanian2023completely}
M.~A. Darmanian, M.~X. Chua, and L.~Wu, ``A completely portable and
  concealable, lightweight assistive exosuit for upper limbs,'' in \emph{2023
  45th Annual International Conference of the IEEE Engineering in Medicine \&
  Biology Society (EMBC)}.\hskip 1em plus 0.5em minus 0.4em\relax IEEE, 2023,
  pp. 1--4.

\bibitem{cai2020real}
S.~Cai, G.~Li, E.~Su, X.~Wei, S.~Huang, K.~Ma, H.~Zheng, and L.~Xie,
  ``Real-time detection of compensatory patterns in patients with stroke to
  reduce compensation during robotic rehabilitation therapy,'' \emph{IEEE
  journal of biomedical and health informatics}, vol.~24, no.~9, pp.
  2630--2638, 2020.

\bibitem{Durandau}
G.~Durandau, W.~F. Rampeltshammer, H.~v.~d. Kooij, and M.~Sartori,
  ``Neuromechanical model-based adaptive control of bilateral ankle
  exoskeletons: Biological joint torque and electromyogram reduction across
  walking conditions,'' \emph{IEEE Transactions on Robotics}, vol.~38, no.~3,
  pp. 1380--1394, 2022.

\bibitem{zaltieri2022assessment}
M.~Zaltieri, D.~L. Presti, M.~Bravi, M.~A. Caponero, S.~Sterzi, E.~Schena, and
  C.~Massaroni, ``Assessment of a multi-sensor fbg-based wearable system in
  sitting postures recognition and respiratory rate evaluation of office
  workers,'' \emph{IEEE Transactions on Biomedical Engineering}, vol.~70,
  no.~5, pp. 1673--1682, 2022.

\bibitem{he2023analysis}
J.~He, D.~Liu, M.~Hou, A.~Luo, S.~Wang, and Y.~Ma, ``Analysis of inter-joint
  coordination during the sit-to-stand and stand-to-sit tasks in stroke
  patients with hemiplegia,'' 2023.

\bibitem{sugai2023lstm}
R.~Sugai, S.~Maeda, R.~Shibuya, Y.~Sekiguchi, S.-I. Izumi, M.~Hayashibe, and
  D.~Owaki, ``Lstm network-based estimation of ground reaction forces during
  walking in stroke patients using markerless motion capture system,''
  \emph{IEEE Transactions on Medical Robotics and Bionics}, 2023.

\bibitem{zhao2019biomechanical}
X.~Zhao, S.~Li \emph{et~al.}, ``A biomechanical analysis of lower limb movement
  on the backcourt forehand clear stroke among badminton players of different
  levels,'' \emph{Applied Bionics and Biomechanics}, vol. 2019, 2019.

\bibitem{nguyen2021quantification}
G.~Nguyen, J.~Maclean, and L.~Stirling, ``Quantification of compensatory torso
  motion in post-stroke patients using wearable inertial measurement units,''
  \emph{IEEE Sensors Journal}, vol.~21, no.~13, pp. 15\,349--15\,360, 2021.

\bibitem{zhao2023analysis}
H.~Zhao, H.~Xu, Z.~Wang, L.~Wang, S.~Qiu, D.~Peng, J.~Li, and J.~Jiang,
  ``Analysis and evaluation of hemiplegic gait based on wearable sensor
  network,'' \emph{Information Fusion}, vol.~90, pp. 382--391, 2023.

\bibitem{oubre2020simple}
B.~Oubre, J.-F. Daneault, K.~Boyer, J.~H. Kim, M.~Jasim, P.~Bonato, and S.~I.
  Lee, ``A simple low-cost wearable sensor for long-term ambulatory monitoring
  of knee joint kinematics,'' \emph{IEEE Transactions on Biomedical
  Engineering}, vol.~67, no.~12, pp. 3483--3490, 2020.

\bibitem{eizentals2020smart}
P.~Eizentals, A.~Katashev, A.~Oks, and G.~Semjonova, ``Smart shirt system for
  compensatory movement retraining assistance: Feasibility study,''
  \emph{Health and Technology}, vol.~10, pp. 861--874, 2020.

\bibitem{cai2019automatic}
S.~Cai, G.~Li, S.~Huang, H.~Zheng, and L.~Xie, ``Automatic detection of
  compensatory movement patterns by a pressure distribution mattress using
  machine learning methods: a pilot study,'' \emph{IEEE Access}, vol.~7, pp.
  80\,300--80\,309, 2019.

\bibitem{liu2021muscle}
X.~Liu, B.~Yang, T.~Liang, J.~Li, C.~Lou, H.~Wang, and X.~Liu, ``Muscle
  compensation analysis during motion based on muscle functional network,''
  \emph{IEEE Sensors Journal}, vol.~22, no.~3, pp. 2370--2378, 2021.

\bibitem{afzal2022evaluation}
T.~Afzal, F.~Zhu, S.-C. Tseng, J.~A. Lincoln, G.~E. Francisco, H.~Su, and S.-H.
  Chang, ``Evaluation of muscle synergy during exoskeleton-assisted walking in
  persons with multiple sclerosis,'' \emph{IEEE Transactions on Biomedical
  Engineering}, vol.~69, no.~10, pp. 3265--3274, 2022.

\bibitem{asemi2022handwritten}
A.~Asemi, K.~Maghooli, F.~N. Rahatabad, and H.~Azadeh, ``Handwritten signatures
  verification based on arm and hand muscles synergy,'' \emph{Biomedical Signal
  Processing and Control}, vol.~76, p. 103697, 2022.

\bibitem{hajiloo2020effects}
B.~Hajiloo, M.~Anbarian, H.~Esmaeili, and M.~Mirzapour, ``The effects of
  fatigue on synergy of selected lower limb muscles during running,''
  \emph{Journal of Biomechanics}, vol. 103, p. 109692, 2020.

\bibitem{ghislieri2020muscle}
M.~Ghislieri, M.~Knaflitz, L.~Labanca, G.~Barone, L.~Bragonzoni, M.~G.
  Benedetti, and V.~Agostini, ``Muscle synergy assessment during single-leg
  stance,'' \emph{IEEE Transactions on Neural Systems and Rehabilitation
  Engineering}, vol.~28, no.~12, pp. 2914--2922, 2020.

\bibitem{kubota2020usefulness}
K.~Kubota, H.~Hanawa, M.~Yokoyama, S.~Kita, K.~Hirata, T.~Fujino, T.~Kokubun,
  T.~Ishibashi, and N.~Kanemura, ``Usefulness of muscle synergy analysis in
  individuals with knee osteoarthritis during gait,'' \emph{IEEE Transactions
  on Neural Systems and Rehabilitation Engineering}, vol.~29, pp. 239--248,
  2020.

\bibitem{liu2022joint}
Y.-X. Liu and E.~M. Gutierrez-Farewik, ``Joint kinematics, kinetics and muscle
  synergy patterns during transitions between locomotion modes,'' \emph{IEEE
  Transactions on Biomedical Engineering}, vol.~70, no.~3, pp. 1062--1071,
  2022.

\bibitem{sheng2021metric}
Y.~Sheng, J.~Zeng, J.~Liu, and H.~Liu, ``Metric-based muscle synergy
  consistency for upper limb motor functions,'' \emph{IEEE Transactions on
  Instrumentation and Measurement}, vol.~71, pp. 1--11, 2021.

\bibitem{yang2017muscle}
N.~Yang, Q.~An, H.~Yamakawa, Y.~Tamura, A.~Yamashita, and H.~Asama, ``Muscle
  synergy structure using different strategies in human standing-up motion,''
  \emph{Advanced Robotics}, vol.~31, no. 1-2, pp. 40--54, 2017.

\bibitem{dupuis2021fatigue}
F.~Dupuis, G.~Sole, C.~Wassinger, M.~Bielmann, L.~J. Bouyer, and J.-S. Roy,
  ``Fatigue, induced via repetitive upper-limb motor tasks, influences trunk
  and shoulder kinematics during an upper limb reaching task in a virtual
  reality environment,'' \emph{PloS one}, vol.~16, no.~4, p. e0249403, 2021.

\bibitem{thomas2023effects}
S.~J. Thomas, G.~C. Castillo, M.~Topley, and R.~W. Paul, ``The effects of
  fatigue on muscle synergies in the shoulders of baseball players,''
  \emph{Sports Health}, vol.~15, no.~2, pp. 282--289, 2023.

\bibitem{matsunaga2021muscle}
N.~Matsunaga, Y.~Okubo, S.~Isagawa, J.~Niitsuma, T.~Otsudo, Y.~Sawada, and
  K.~Akasaka, ``Muscle fatigue in the gluteus maximus changes muscle synergies
  during single-leg landing,'' \emph{Journal of Bodywork and Movement
  Therapies}, vol.~27, pp. 493--499, 2021.

\bibitem{baechle2008essentials}
T.~R. Baechle and R.~W. Earle, \emph{Essentials of strength training and
  conditioning}.\hskip 1em plus 0.5em minus 0.4em\relax Human kinetics, 2008.

\bibitem{kennedy2014methods}
C.~Kennedy-Armbruster and M.~Yoke, \emph{Methods of group exercise
  instruction}.\hskip 1em plus 0.5em minus 0.4em\relax Human Kinetics, 2014.

\bibitem{vourganas2019factors}
I.~Vourganas, V.~Stankovic, L.~Stankovic, A.~Kerr \emph{et~al.}, ``Factors that
  contribute to the use of stroke self-rehabilitation technologies: A review,''
  \emph{JMIR Biomedical Engineering}, vol.~4, no.~1, p. e13732, 2019.

\bibitem{UpperLimbModelVICON}
V.~M.~S. Lmt, ``Upper limb model,'' Available at
  \url{https://www.vicon.com/software/models-and-scripts/upper-limb-model/}
  (1/7/2007).

\bibitem{Seniam}
P.~R. Merletti, ``Seniam,'' Available at \url{http://seniam.org/}.

\bibitem{ABC}
P.~Konard, ``The abc of emg,'' Available at
  \url{https://www.noraxon.com/wp-content/uploads/2014/12/ABC-EMG-ISBN.pdf}
  (4/3/2006).

\bibitem{turpin2021improve}
N.~A. Turpin, S.~Uriac, and G.~Dalleau, ``How to improve the muscle synergy
  analysis methodology?'' \emph{European journal of applied physiology}, vol.
  121, no.~4, pp. 1009--1025, 2021.

\bibitem{boyas2009changes}
S.~Boyas, O.~Ma{\"\i}setti, and A.~Gu{\'e}vel, ``Changes in semg parameters
  among trunk and thigh muscles during a fatiguing bilateral isometric
  multi-joint task in trained and untrained subjects,'' \emph{Journal of
  Electromyography and Kinesiology}, vol.~19, no.~2, pp. 259--268, 2009.

\bibitem{wang2022antagonist}
L.~Wang, X.~Song, H.~Yang, C.~Wang, Q.~Shao, H.~Tao, M.~Qiao, W.~Niu, and
  X.~Liu, ``Are the antagonist muscle fatigued during a prolonged isometric
  fatiguing elbow flexion at very low forces for young adults?''
  \emph{Frontiers in Physiology}, vol.~13, p. 956639, 2022.

\bibitem{guo2021human}
L.~Guo, Z.~Lu, and L.~Yao, ``Human-machine interaction sensing technology based
  on hand gesture recognition: A review,'' \emph{IEEE Transactions on
  Human-Machine Systems}, vol.~51, no.~4, pp. 300--309, 2021.

\end{thebibliography}

\end{document}